\pdfoutput=1

\documentclass[11pt]{article}

\usepackage[final]{acl}

\usepackage{times}
\usepackage{latexsym}

\usepackage[T1]{fontenc}

\usepackage[utf8]{inputenc}

\usepackage{microtype}

\usepackage{inconsolata}

\usepackage{graphicx}

\usepackage{amssymb}

\title{Llamazip: Leveraging LLaMA for Lossless Text Compression and Training Dataset Detection}

\author{Sören Dréano \\
  ML-Labs \\
  Dublin City University \\
  \texttt{soren.dreano2@mail.dcu.ie} \\\And
  Derek Molloy \\
  School of Electronic Engineering \\
  Dublin City University \\
  \texttt{derek.molloy@dcu.ie}
  \AND
  Noel Murphy \\
  School of Electronic Engineering \\
  Dublin City University \\
  \texttt{noel.murphy@dcu.ie} \\}

\begin{document}
\maketitle
\begin{abstract}
This work introduces Llamazip, a novel lossless text compression algorithm based on the predictive capabilities of the LLaMA3 language model. Llamazip achieves significant data reduction by only storing tokens that the model fails to predict, optimizing storage efficiency without compromising data integrity. Key factors affecting its performance, including quantization and context window size, are analyzed, revealing their impact on compression ratios and computational requirements.

Beyond compression, Llamazip demonstrates the potential to identify whether a document was part of the training dataset of a language model. This capability addresses critical concerns about data provenance, intellectual property, and transparency in language model training.
\end{abstract}

\section{Introduction}

Lossless compression is, by definition, a completely reversible operation. One of its main use cases is in archival and backup, since the objective is to be able to recreate an exact copy of the data at a specific time while saving on storage costs compared to a non-compressed archive. Lossless compression is also one of the cornerstones of web performance. In fact, more than 88\% websites rely on lossless text compression to reduce required bandwidth and accelerate loading times \cite{compression_stats}.

Lossless compression is a trade-off between savings on data storage and bandwidth, since compressed data require less space, and the computing time needed to recover the data, since instead of merely accessing the data it is first necessary to decompress it. However, modern hardware and software algorithms are highly efficient, and the decompression time is often negligible compared to the benefits.

The compression ratio is a measure of the efficiency of a compression algorithm. It expresses the ratio between the size of the data before compression and the size of the data after compression. The focus of this research is on compression efficiency, and the speed of compression and decompression is beyond the scope of the analysis presented here. The efficiency, as measured by the compression ratio, of the different compression algorithms varies considerably depending on the nature of the data to be compressed \cite{8117850}. The compression ratio formula is as follows:

\begin{center}
$ratio = \frac{Size\_original\_data}{Size\_compressed\_data}$
\end{center}

\section{Next-Token Prediction}

In a character string, the probability that a continuous sequence of characters will be repeated increases with the size of the string. This is particularly likely if the character string represents natural languages, where repetitions of words and roots are frequent. The LZMA \cite{lzma} compression algorithm, which is an improvement over LZ77, features a larger sliding window along with models based on Markov chains to optimize the compression ratio, since it relies on continuous characters to predict upcoming symbols, based on previous sequences. It also relies on range coding to compress the ranges and lengths of the pointers, further optimizing the compression ratio. 

In a similar manner to the way that Markov chains can be used in lossless text compression, since they compress more efficiently by predicting the next characters, recent advances in Machine Learning have also greatly improved the predictive capabilities of the models.

Next-Token Prediction (NTP) is the task of predicting the next token based on all or some of the previous tokens. Although NTP is usually employed as a form of pre-training in many Natural Language Processing (NLP) models such as GPT, it can also be used directly for text compression. nncp \cite{nncp} is a collection of LSTM and Transformer models that have been trained on NTP to perform lossless text compression. nncp is the current state-of-the-art on the Large Text Compression Benchmark, as mentioned in Section \ref{sec:enwik9}.

\section{Datasets}

To evaluate the performance of our lossless text compression algorithms, a diverse set of test documents was gathered as described below. The documents are plain text files containing English text. An important aspect to consider is the publication date of the files. Since our work relies on a pre-trained model, that was trained on a large corpora of text, it is crucial to consider whether the test document could have been part of the training dataset of the underlying model.

\subsection{Wikipedia}
\label{sec:enwik9}

The Large Text Compression Benchmark \cite{large_text_compression_benchmark} is a competition that has been running since 2006, with the goal of obtaining the best compression ratio on the first $10^9$ bytes of the English Wikipedia dump on March 3, 2006, also known as enwik9. This dataset contains mainly English text; therefore, the results of this benchmark are not fully representative of the compression ratios for text in other languages.

\subsection{Alice29}
\label{sec:alice29}

The Canterbury corpus \cite{arnold1997corpus} is a collection of files of different types, used to benchmark lossless compression algorithms. The size of the corpus is 2,810,784 bytes and includes the file ``alice29.txt'', which is a text version of the well-known novel, ``Alice's Adventures in Wonderland'' from Lewis Carroll, which comprises 148,479 bytes. The Squash Compression Benchmark \cite{squash_benchmark} maintains a leaderboard for compression algorithms on a collection of files, including alice29.txt.

\subsection{Frankenstein}

Frankenstein is an iconic novel from Mary Shelley published in 1818 in which Victor Frankenstein, a young scientist, creates a living being by assembling and reanimating body parts. The file was obtained from Codebreaker Github repository \cite{codebreaker}, which was first obtained from Project Gutenberg. The original file contains 441,034 bytes.

\subsection{FSDFSD}

Archive of Our Own \cite{archiveofourown}, also known as AO3, is a ``fan-created, fan-run, nonprofit, noncommercial archive for transformative fanworks, like fanfiction, fanart, fan videos, and podfic''. AO3 provides a search engine with a wide range of criteria. FSDFSD \cite{fsdfsd_ao3} is a novel published on AO3 by user ``Scarpoile'' after the release of LLaMA3, which has been marked as not containing content marked as ``Not Safe For Work'' and is in English. Its size of 1,253,946 bytes was shortened to 441,034 bytes.

\subsection{LLaMAgen}

``Meta-Llama-3-8B-Instruct'' is a Large Language Model (LLM), fine-tuned for dialogue and answering questions. The document was generated using two prompts, namely the system prompt, which remained constant during the generation, and the user prompt. The generation parameters are detailed in Appendix \ref{sec:generation_parameters}. The file is named ``LLaMAgen''.

\subsection{Mistralgen}

``Mistral-7B-Instruct-v0.3'' \cite{jiang2023mistral7b} is another LLM, fine-tuned for dialogue and question answering. The generation parameters are detailed in Appendix \ref{sec:generation_parameters}. The generated file contains 565,146 bytes and is named ``Mistralgen''.

\subsection{MIT books}

At the time the present research was being carried out, two out of the most recent open access books published are \textit{The Aesthetics of Stealth: Digital Culture, Video Games, and the Politics of Perception} \cite{10.7551/mitpress/15189.001.0001} and \textit{Reimagining the More-Than-Human City: Stories from Singapore} \cite{10.7551/mitpress/15539.001.0001}, titles that will be abbreviated as MIT1 and MIT2, respectively. For the purposes of this study, it is assumed that the authors of these books were not assisted in their writing by LLMs. The Python library pypdf \cite{pypdf} was used to retrieve the text from the PDF files of the books, by applying the ``extract\_text()'' function to all the pages and concatenating the strings. After conversion to text only, MIT1 contains 622,101 bytes and MIT2 699,783 bytes.

\section{Llamazip}

LLaMA3 \cite{grattafiori2024llama3herdmodels} is a recently released Open Source Large LLM that relies on NTP for pre-training. Llamazip is an LLaMA3-built lossless text compression algorithm dedicated to data archival and storage, as it trades compression and decompression speed for maximal reduction in the size of the data. Using the predictive capabilities of LLaMA3, Llamazip only stores mistaken predictions to minimize storage requirements. Llamazip relies on the 8 billion parameter version of LLaMA3.

The work presented in this Section also examines the effects of LLM quantization and context window size on the performance of the NTP task. Quantization is a technique that reduces the size of the parameters and can potentially speed up both inference and training by simplifying calculations, although it can also result in a decrease in the precision of the parameters. The context window refers to the number of tokens considered when predicting the next token; a larger window requires more memory and longer computation times. Analyzing the impact of these two variables reveals performance variations in Llamazip, thereby reflecting the performance of the underlying model.

\subsection{Options}

The model has several options that affect the efficiency of the algorithm, measured by the compression ratio. The exact parameters used to compress a file must be known at the decompression time to retrieve the original data.

\begin{itemize}
    \item Quantization: Recent advances in quantization allow for a reduction in the memory requirement of large models, at the cost of a decrease in the compression ratio. Different quantizations are compared: float16, bfloat16, int8 and int4, each representing different levels of precision. Float32 (no quantization) was not tested, as Llama3 cannot fit on the available hardware.
    \item Maximum Window Size: Although the compression ratio would probably be maximized by using all previous tokens to predict the next token, the size of the available memory restricts the number of tokens that can be exploited for NTP. Although the window size can take any value in $ \mathbb{N}^*$, it was restricted to the set [16, 32, 64, 128, 256, 512, 1024, 2048] in the context of the current work, since each additional token increases the memory requirements of the algorithm.
\end{itemize}

\subsection{Algorithm}

In this section, we present a detailed explanation of Llamazip, a lossless text compression algorithm. The algorithm is designed to efficiently reduce the size of a text file, while ensuring that the original data can be fully recovered without any loss of information.

\subsubsection{Compression}

Compression relies mainly on LLaMA3's predictive capacity, since once a token is correctly predicted, there is no need to record it, and therefore there is only a need to record errors.

\begin{enumerate}
    \item The counter is initialised with the value 0.
    \item The text to compress is tokenized.
    \item The first 10 tokens are added to a list and saved directly at the beginning of the output file.
    \item The model tries to predict the next token from the tokens stored in the list until the file is fully compressed. There are two possibilities:
    \begin{itemize}
        \item If the prediction is successful, the value of the counter is incremented, and the predicted token is appended to the list.
        \item If the prediction is wrong, the counter is recorded in the output file and then set to 0. The index of the token, preceded by the character ``@'' to distinguish it from the value of the counter, is recorded in the output file and appended to the list.
    \end{itemize}
    \item After each prediction, the size of the list is examined, and once again there are two possibilities:
    \begin{itemize}
        \item If the size of the list is less than the maximum window size, no further operations are required.
        \item If the size of the list is equal to the maximum window size, the oldest value of the list is discarded.
    \end{itemize}
    \item Once all the data have been read, the counter value is written for the last time to the end of the file.
    \item The file is then further compressed using gzip.
\end{enumerate}

\subsubsection{Decompression}

Decompression involves following the pattern recorded during compression, i.e. alternating between prediction sequences and recovery sequences for tokens written as they are in the file.

\begin{enumerate}
    \item The file is first decompressed using gzip.
    \item The first 10 tokens are added to a list.
    \item The next value is read. There are two possibilities:
    \begin{itemize}
        \item If the first character of the value is ``@'', it means that it is an index of a token and not a counter value. The index of the token is appended to the list and to an intermediate file. If the size of the list corresponds to the maximum window size, the oldest token is discarded from the list.
        \item Else, the LLaMA3 model predicts $n$ tokens, where $n$ is the value of the counter. After each prediction, the token is added to the list and to the intermediate file, and if required, the oldest token is removed from the beginning of the list.
    \end{itemize}
    \item The intermediate file is then detokenized in order to retrieve the characters.
\end{enumerate}

\subsection{Similar Work}

LLMZip \cite{llmzip} is a compression algorithm that relies on LLaMA1, in its 7 billion parameter version, to compress text data. Similarly to Llamazip, LLMZip relies on the NTP capabilities of a pre-trained LLM to avoid recording every token. However, while Llamazip only registers tokens that were accurately predicted and discards inaccurate hypotheses, LLMZip saves the rank of the correct token. In fact, at each step, an LLM can output the probability that each token in the vocabulary is the next token, so it is possible to sort the tokens by probability and record the position of the correct token. In this manner, LLMZip only records integers, where the lowest values mean that the tokens are among the most likely and the highest values mean that the LLM estimated that the token only had a low probability of appearing at this location. 

\section{Results}

Various text files, all in English, have been compressed using Llamazip. Each file was compressed with 8 different window sizes, as well as 4 different quantization settings, for a total of 32 combinations per file. The sizes and compressions ratios between the optimal Llamazip parameters and different compression algorithms are then compared.

\subsection{Effects of Quantization and Context Window}

\begin{figure} [t]
	\centering
	\includegraphics[width=\columnwidth]{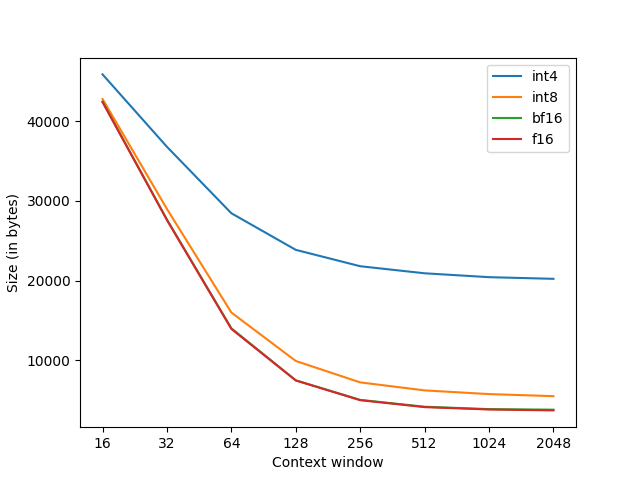}
	\caption{Compression of alice29 depending on the context window and the quantization method}
	\label{fig:alice29_compression}
\end{figure}

Figure \ref{fig:alice29_compression} shows the size of the novel compressed by Llamazip depending on the parameters used. It is immediately apparent that a long context window and quantization that only halves the initial size of the model are factors that improve the compression ratio. Both float16 and bfloat16 display similar efficiencies.

\begin{figure} [t]
	\centering
	\includegraphics[width=\columnwidth]{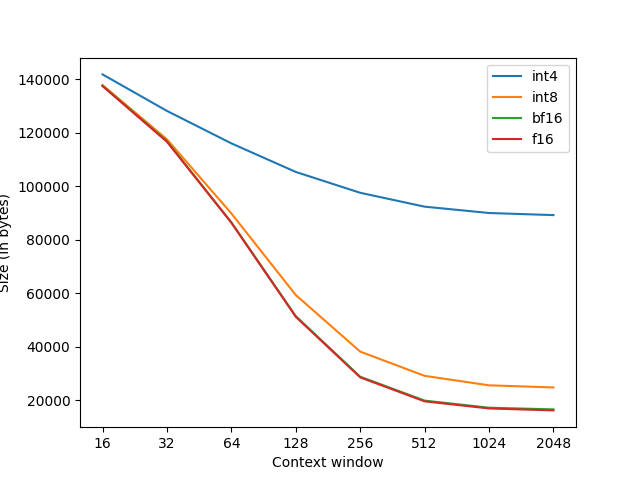}
	\caption{Compression of Frankenstein depending on the context window and the quantization method}
	\label{fig:frankenstein_compression}
\end{figure}

Figure \ref{fig:frankenstein_compression} shows the evolution of the size of the Frankenstein novel depending on the quantization and context window. Similarly to the alice29.txt file, the length of the context window and the quantization method have a large impact on the size of the compressed file.

\begin{figure} [t]
	\centering
	\includegraphics[width=\columnwidth]{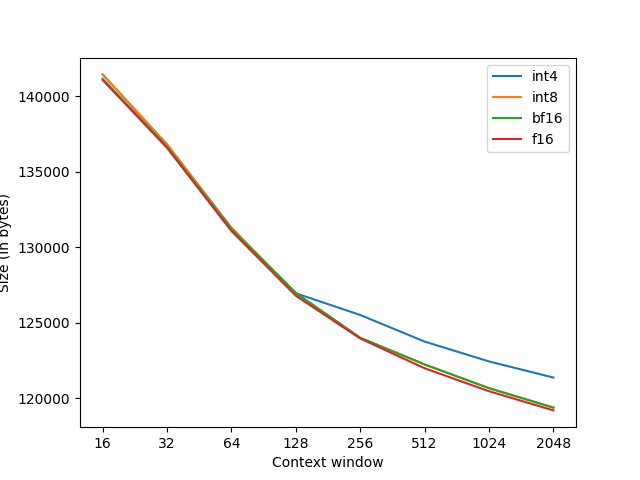}
	\caption{Compression of FSDFSD depending on the context window and the quantization method}
	\label{fig:FSDFSD_compression}
\end{figure}

The excellent compression ratios of Llamazip for both alice29 and Frankenstein documents are probably due to the fact that these documents were part of the dataset on which the underlying model, LLaMA3, was trained. It is crucial to check the effectiveness of Llamazip on a document more recent than the LLaMA3 release date, as this is the only way to ensure that the document is not part of the training set. Following the same pattern as previously observed, Figure \ref{fig:FSDFSD_compression} demonstrates the impact the context window and quantization method have. 

\begin{figure} [t]
	\centering
	\includegraphics[width=\columnwidth]{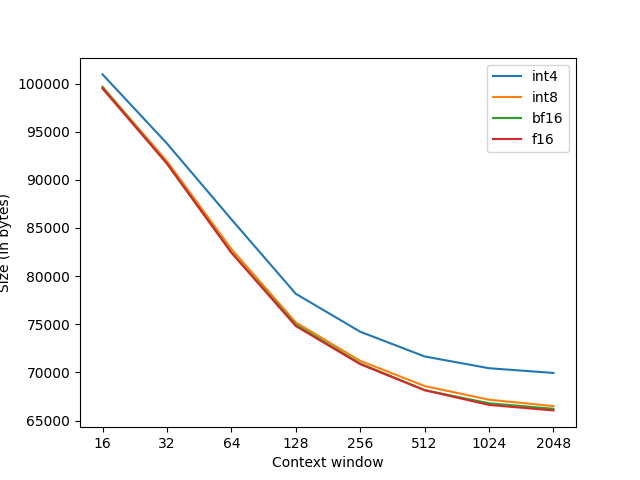}
	\caption{Compression of the LLaMAgen file depending on the context window and the quantization method}
	\label{fig:generated_llama_compression}
\end{figure}

Consistently with previous results, Figure \ref{fig:generated_llama_compression} shows how a longer compression window with less lossy quantization outperforms other parameters for the LLaMAgen file.

\begin{figure} [t]
	\centering
	\includegraphics[width=\columnwidth]{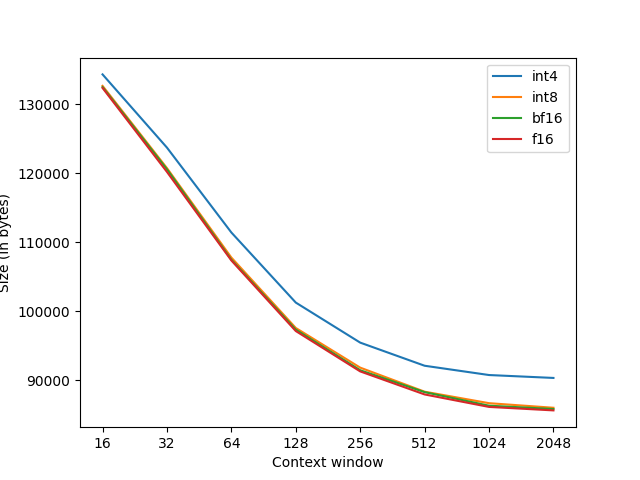}
	\caption{Compression of the Mistralgen file depending on the context window and the quantization method}
	\label{fig:generated_mistral_compression}
\end{figure}

The results obtained for the Mistralgen document are similar to those obtained on LLaMAgen and are shown in Figure \ref{fig:generated_mistral_compression}.

\begin{figure} [t]
	\centering
	\includegraphics[width=\columnwidth]{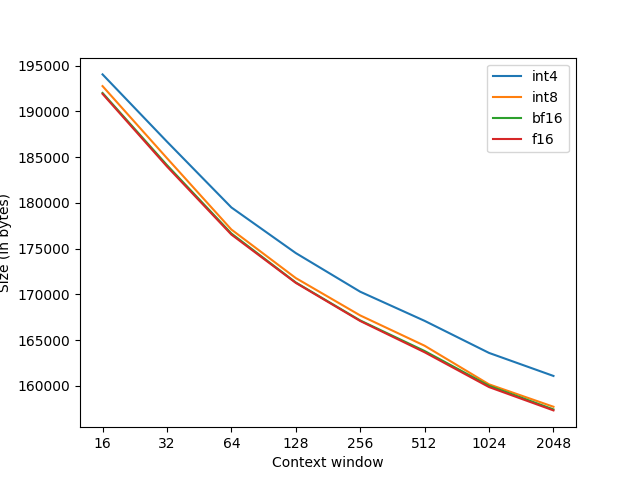}
	\caption{Compression of the MIT1 file depending on the context window and the quantization method}
	\label{fig:book_9780262380768_compression}
\end{figure}

\begin{figure} [t]
	\centering
	\includegraphics[width=\columnwidth]{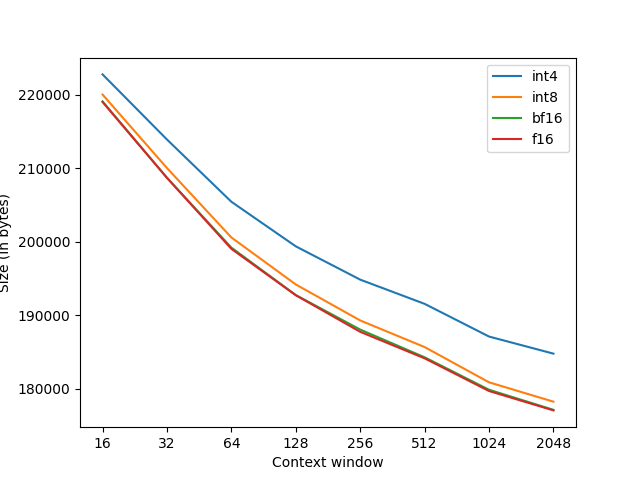}
	\caption{Compression of the MIT2 file depending on the context window and the quantization method}
	\label{fig:book_9780262381406_compression}
\end{figure}

Figures \ref{fig:book_9780262380768_compression} and \ref{fig:book_9780262381406_compression} show the compressed sizes of the MIT1 and MIT2 documents depending on the context window and quantization. The impact of these settings is less significant than for the alice29 and Frankenstein documents.

In line with the authors' expectations, the compression ratio is positively linked to a longer context window, enabling the next token to be predicted more accurately, thus reducing the number of tokens that need to be hard-coded in the file. The compression ratio is also correlated with a low-loss quantization, which reduces the performance of the underlying model less than a very lossy quantization such as int4.

\subsection{Wikipedia}

As Llamazip is an expensive algorithm and since the size of enwik9 is 1 GiB, it would unfortunately take too long to compress it in its entirety. Using the LLaMA3 tokenizer, the enwik9 dataset is 288,204,082 tokens long, which is quite substantially more than the 300,972 of the novel FSDFSD. Thus, similarly to LLMZip \cite{llmzip}, the performance of Llamazip is measured over 10 batches of 100,000 tokens each, which are sampled uniformly on the 13,147,026 lines of the dataset. Table \ref{tab:llamazip_enwik9} in Appendix \ref{sec:tables} presents the compression ratios of the different batches of 100,000 tokens.

\subsection{Comparison with other Algorithms}

Only the compression ratio of the document is compared between the different compression algorithms, while the computation time required or the size of the model are outside the scope of this work. In addition, Llamazip is presented here with its best-performing options, namely a context window of 2048 tokens and float16 quantization. This section evaluates its effectiveness by comparing the compression ratios achieved by Llamazip against commonly used methods such as Brotli, xz, zstd, gzip, zpaq, and nncp. Table \ref{tab:llamazip_ratio} shows that Llamazip consistently achieves higher compression ratios across datasets.

\begin{table*}[t]
    \centering
    \caption{Compression ratio depending on the compression algorithm}
    \scalebox{1.0}{
    \begin{tabular}{c|c|c|c|c|c|c|c}
         Document& Llamazip& brotli& xz& zstd& gzip& zpaq& nncp\\
         \hline
         alice29& \textbf{40.32}& 3.23& 3.10&2.64 & 2.78& 2.27& 2.02\\
         Frankenstein& \textbf{27.10}& 3.20& 3.11& 2.68& 2.64& 2.34& 2.27\\
         FSDFSD& \textbf{3.77}& 3.43& 3.34& 2.84& 2.80& 2.51& 2.33\\
         LLaMA& \textbf{6.68}& 4.99& 4.88& 3.74& 3.60& 3.45& 2.53\\
         Mistral& \textbf{6.60}& 4.69& 4.61& 3.54& 3.43& 3.43& 2.59\\
         MIT1& \textbf{3.95}& 3.56& 3.43& 2.90& 2.85& 2.60& 2.32\\
         MIT2& \textbf{3.95}& 3.60& 3.46& 2.91& 2.82& 2.60& 2.39\\
    \end{tabular}
    }
    \label{tab:llamazip_ratio}
\end{table*}

Table \ref{tab:llamazip_compressed_size} in Appendix \ref{sec:tables} provides the sizes of the compressed files depending on the compression algorithm.

The \textit{bits per character} metric is a fundamental measure in text compression as it provides a straightforward way to evaluate the efficiency of a compression method. This metric quantifies the average number of bits required to represent each character in the compressed text, providing a clear and interpretable indicator of compression performance. This metric is particularly valuable for comparing methods across datasets of varying lengths and content, as it normalises the compression efficiency relative to the input size.

Table \ref{tab:bits_per_char} (bpc) in Appendix \ref{sec:tables} displays the number of bits per character required by different text compression algorithms depending on the document. Concerning the enwik9 document, the 10 batches of 100,000 tokens each are concatenated. 

FSDFSD, MIT1 and MIT2 are the three documents that were published after the release date of LLaMA3 and thus, are known to not be in the training set. On average, Llamazip requires 2.0693 bits per character. On a document that is likely not part of the underlying LLaMA training dataset, the original authors of LLMZip found that their algorithm compresses a document using 0.8426 bits per character.

This shows that recording the rank of the prediction rather than recording the token itself if the prediction is wrong is a more effective strategy for compressing the data. However, the work done on LLMZip has focused solely on text compression, whereas the present study finds that Llamazip might detect if a document is part of the training set of an LLM.

\subsection{Speed and Memory Requirements}

Table \ref{tab:token_per_second} in Appendix \ref{sec:tables} displays the number of tokens compressed per second depending on the quantization and context window. The compression speed is perfectly negatively correlated with the window size, with a Spearman's $\rho$ of -1.

Modern hardware has dedicated float32 accelerators, which can even perform two float16 operations in the same time frame required by a single float32 operation. However, the NVidia 4090, which at the time of writing is the highest-end NVidia consumer-grade Deep Learning accelerator, does not have any int4 or int8 accelerators. Llamazip compresses data at least as fast using float16 and bfloat16 quantizations compared with integer quantizations at equal context window sizes, while offering a better compression ratio.

The memory requirements displayed in Table \ref{tab:memory_requirements} in Appendix \ref{sec:tables} are based on the memory usage obtained using the command ``nvidia-smi'', an official program provided by NVidia. Consistent with expectations, both quantizations and the size of the context window have a positive impact on the quantity of memory used during compression, since an int4 parameter requires 4 times fewer bits than a float16 one and the context window has to be entirely kept in memory to predict the next token. As expected, bfloat16 and float16 require the exact same amount of memory, since both formats use 16 bits per parameter. Each larger quantization size, even with the smallest context window, requires more memory than the smaller size with the longest context window.

\section{Identifying Documents in the Training Dataset}
\label{sec:llamazip_training_set_detection}

The performance of the Llamazip algorithm varies widely depending on the document being compressed, with previously seen data being compressed more than data that are outside of the training set.

The hypothesis proposed here is that the observed difference in compression ratios between the highest-performing and lowest-performing parameters of Llamazip, depending on whether the document is present in the underlying training set, is attributed to the model’s memorisation of the document. By intentionally limiting the model's capacity to recall its training set, through a reduction in parameter precision, its access to memory is diminished. As a result, when the memory of the training set has little or no influence on compressing new data, the difference in compression ratios is minimal. Conversely, when the data have been previously encountered and are stored in memory, this reduction in capacity leads to a significant difference in compression ratios.

Two key indicators emerge:

\begin{itemize}
    \item As demonstrated in Table \ref{tab:worst_best}, the ratio between the size of the file compressed with the worse Llamazip parameters, which are a int4 quantisation along a context window of only 16 tokens, and the best parameters, which consist in float16 quantisation and a context window of 2048 tokens, is larger on samples from the training set;
    \item As shown in Table \ref{tab:llamazip_ratio}, the difference in the compression ratio obtained by Llamazip compared to the other algorithms is much larger on documents belonging to the LLaMA3 dataset.
\end{itemize}

\begin{table*}[t]
    \centering
    \caption{Variations in the size of the compressed documents depending on the quantization and context window of Llamazip}
    \scalebox{1.0}{
    \begin{tabular}{c|c|c|c}
         Document& Largest size (bytes)& Smallest size (bytes)& Ratio between largest and smallest\\
         \hline
         alice29& 45,911& 3,682& 12.47\\
         Frankenstein& 141,876& 16,273& 8.72\\
         FSDFSD& 141,446& 119,197& 1.19\\
         LLaMAgen& 100,948& 66,055& 1.53\\
         Mistralgen& 134,335& 85,632& 1.57\\
    \end{tabular}
    }
    \label{tab:worst_best}
\end{table*}

\subsection{Source code}

The source code of Llamazip is available at [anonymized].

\section{Future Work}

The unquantized LLaMA3 model has not been tested due to hardware limitations. Further work is required to inspect the performance impact of using float32 weights.

Llamazip should register out-of-vocabulary tokens as they are, rather than registering the index as for other tokens, which would make the code only slightly more complex. This would mitigate any possible data loss due to the replacement of out-of-vocabulary tokens by \textit{<unk>} tokens.

As the name suggests, Llamazip is an algorithm based on the LLaMA family of models. Other open-source models might obtain better compression performances and open the door to a larger language diversity.

Llamazip could adopt the strategy used by a previous work, LLMZip, which is to save the rank of the correct token rather than saving the correct token in the event of a bad prediction.

As demonstrated in Section \ref{sec:llamazip_training_set_detection}, it is possible that Llamazip can be used to detect whether a document was present in the original dataset of an LLM, at least in the case where it was trained on a NTP task. However, since the preliminary work is based on a single document, further work is needed to verify the reproducibility of this hypothesis. However, this task is made more difficult by the relative obscurity of the datasets used to train LLMs and the constant increase in the distribution of machine-generated content, which is not always labelled as such. This last point is of particular importance and projects such as Wordfreq \cite{wordfreq}, which is a database of word frequencies for various natural languages, will not be updated by its author since generative models have ``polluted'' the data, which is no longer representative of human usage. For example, it has been observed \cite{delving_into_delve} that ChatGPT overuses the word ``delve'' compared to humans.

\section{Limitations}

Although Llamazip demonstrates promising results in lossless text compression and dataset provenance detection, it has several limitations that must be acknowledged:

\subsection{Uncertainty Regarding Training Dataset}

The effectiveness of Llamazip in compressing certain documents may be influenced by whether those documents were part of LLaMA3’s original training set. However, due to the lack of transparency in the construction of LLaMA3’s dataset and the computational challenges associated with analyzing such a large-scale corpus, the authors cannot definitively confirm whether texts such as Alice’s Adventures in Wonderland (alice29) and Frankenstein were included in LLaMA3’s training data. This uncertainty introduces potential biases in compression performance that could impact our findings.

\subsection{Potential Contamination in Post-LLaMA3 Texts}

The authors assumes that documents published after the release of LLaMA3 were not included in its training data. However, this assumption is imperfect, as many contemporary texts, even those published after LLaMA3’s release, may have been generated or edited with the assistance of large language models. Given the increasing prevalence of AI-assisted writing, it is possible that some of these documents contain stylistic or linguistic patterns that are already well-represented in LLaMA3’s training data, leading to unexpectedly high compression rates. This complicates the task of identifying whether a document was truly unseen by the model.

\subsection{Language Scope Restriction}

The current study is limited to English-language text, as LLaMA3 was primarily trained on English data. Consequently, Llamazip’s ability to compress non-English text remains unexplored. Since linguistic structures, tokenization schemes, and model memorization behaviors differ across languages, extending Llamazip to multilingual datasets may require significant modifications and further evaluation.

\section{Conclusion}

In conclusion, this work has demonstrated the multifaceted potential of Llamazip as a novel, lossless text compression algorithm that leverages the predictive capacity of LLaMA3. By only storing the tokens that the model fails to predict accurately, Llamazip achieves significant compression, optimising storage requirements without compromising data integrity. This approach to compression trades processing speed for maximal data reduction, making it well-suited for archival purposes where storage efficiency takes precedence over rapid decompression.

Beyond its primary function in data compression, Llamazip introduces an innovative capability to verify whether specific documents were included in the training set of an LLM. By comparing its compression performance under various parameter settings, a discernible pattern emerges that differentiates known documents from previously unseen ones. This property positions Llamazip as a promising tool for identifying the provenance of data within an LLM training context, a crucial capability to address concerns about data privacy, intellectual property, and transparency of training sources. 

Finally, this word has also explored the effects of model quantization and context window size on Llamazip’s performance, highlighting how these variables influence next-token prediction accuracy and, consequently, compression efficiency.

\section{Acknowledgments}

This publication has emanated from research conducted with the financial support of Science Foundation Ireland under Grant number 18/CRT/6183. For the purpose of Open Access, the author has applied a CC BY public copyright licence to any Author Accepted Manuscript version arising from this submission.

\bibliography{custom}

\appendix

\section{Generation Parameters}
\label{sec:generation_parameters}

\begin{itemize}
    \item System prompt: ``You are a novelist chatbot who always write long novels. You are specialized in original fairy tales. You do not commit the wrongful act of plagiarism and invent your own stories. You write long fairy tales.';
    \item Initial user prompt: ``Once upon a time,'';
    \item The following user prompts are the last 50 tokens generated during the last generation.
\end{itemize}

The novel was generated in 160 steps, using a temperature of 0.5 and the model generated at least 256 tokens during each step. Since the seeds of the Random Number Generators were set, the generation is fully reproducible. The final file was shortened to 441,064 bytes.

\section{Tables}
\label{sec:tables}

\begin{table*}[t]
    \centering
    \caption{Size of the different compressed batches of the enwik9 dataset using Llamazip}
    \scalebox{1.0}{
    \begin{tabular}{c|c|c|c}
         Index (in lines)& Size of the batch (bytes)& Size of the compressed batch (bytes)& Compression ratio\\
         \hline
         0& 374295&72811& 5.14 \\
         1,314,702& 361,735&75,118& 4.82 \\
         2,629,404& 247,737&20,719& 11.96 \\
         3,944,106& 353,510&73,806& 4.79 \\
         5,258,808& 327,783&45,000& 7.28 \\
         6,573,510& 334,072&35,649& 9.37 \\
         7,888,212& 368,903&86,412& 4.27 \\
         9,202,914& 340,115&68,056& 5.00 \\
         10,517,616& 345,954&71,151& 4.86 \\
         11,832,318& 361,407&81,016& 4.46 \\
         Total& 3,415,511& 629,738& 5.42\\
    \end{tabular}
    }
    \label{tab:llamazip_enwik9}
\end{table*}

\begin{table*}[t]
    \centering
    \caption{Size (in bytes) of the compressed documents depending on the compression algorithm}
    \scalebox{1.0}{
    \begin{tabular}{c|c|c|c|c|c|c|c}
         Document& Llamazip& brotli& xz& zstd& gzip& zpaq& nncp\\
         \hline
         alice29& \textbf{3,682}& 45,999& 47,872& 56,245 &53,429 & 65,312& 73,472\\
         Frankenstein& \textbf{16,273}& 137,788& 141,644& 164,549& 166,868& 188,821& 194,258\\
         FSDFSD& \textbf{119,197}& 130,780& 134,428& 157,801& 160,121& 179,020& 189,197\\
         LLaMAgen& \textbf{66,055}& 88,409& 90,360& 117,813& 122,395& 127,767& 174,121\\
         Mistralgen& \textbf{85,632}& 120,549& 122,596& 159,709& 164,880& 173,084& 218,203\\
         MIT1& \textbf{157,315}& 174,798& 181,600& 214,394& 217,997& 239,460& 268,320\\ 
         MIT2& \textbf{177,092}& 194,484& 202,032& 240,193& 248,145& 268,747& 292,252\\
    \end{tabular}
    }
    \label{tab:llamazip_compressed_size}
\end{table*}

\begin{table*}[ht]
    \centering
    \caption{Number of bits necessary to represent one character depending on the document and the algorithm}
    \scalebox{0.85}{
    \begin{tabular}{c|c|c|c|c|c|c|c}
         Document& Characters & Llamazip (bpc)& brotli (bpc) & xz (bpc) & zstd (bpc) & zpaq (bpc)& nncp (bpc)\\
         \hline
         alice29& 148,479 & \textbf{0.200} & 2.480 & 2.576 & 3.032 & 3.520& 3.960\\
         Frankenstein& 441,034& \textbf{0.296}& 2.496& 2.568& 2.984& 3.424& 3.528\\
         FSDFSD& 441,034& \textbf{2.160}& 2.368& 2.440& 2.864& 3.248& 3.504\\
         LLaMAgen& 619,631& \textbf{1.184}& 1.528& 1.552& 2.088& 2.240& 2.248\\
         Mistralgen& 565,146& \textbf{1.208}& 1.704& 1.736& 2.264& 2.448& 3.088\\
         MIT1& 622,101& \textbf{2.048}& 2.240& 2.336& 2.760& 3.080& 3.448\\
         MIT2& 699,783& \textbf{2.024}& 2.224& 2.312& 2.744& 3.072& 3.344\\
         enwik9& 3,407,633& \textbf{1.480}& 1.904& 1.936& 2.440& 2.608& 1.88\\
    \end{tabular}
    }
    \label{tab:bits_per_char}
\end{table*}

\begin{table*}[t]
\centering
\caption{Number of tokens compressed per second using Llamazip depending on the context window and the quantization using a NVidia 4090}
\begin{tabular}{l|l|l|l|l|l|l|l|l}
             & \multicolumn{8}{l}{Context window size (in tokens)}      \\
Quantization & 16 & 32 & 64 & 128 & 256 & 512 & 1024 & 2048 \\
\hline
int4         & 10 & 10 &  9 &  8  &  6  &  4  &  2   &  1   \\
int8         & 12 & 12 & 12 & 12  & 12  & 11  &  8   &  5   \\
blfloat16    & 48 & 46 & 44 & 32  & 31  & 18  &  9   &  5   \\
float16      & 49 & 46 & 44 & 39  & 31  & 18  &  9   &  5   \\
\end{tabular}
    \label{tab:token_per_second}
\end{table*}

\begin{table*}[t]
\centering
\caption{Size (in MiB) of the memory required to compress a file using Llamazip depending on the context window and the quantization}
\scalebox{1.0}{
\begin{tabular}{l|l|l|l|l|l|l|l|l}
             & \multicolumn{8}{l}{Context window size (in tokens)}      \\
Quantization & 16 & 32 & 64 & 128 & 256 & 512 & 1024 & 2048 \\
\hline
int4       &6,751 &6,753 &6,759 &6,769 &6,781 &6,815 &7,037 &7,485 \\
int8       &9,275 &9,277 &9,285 &9,297 &9,313 &9,395 &9,881 &12,321\\
blfloat16  &15,943&15,945&15,949&15,961&15,993&16,077&16,485&18,291\\
float16    &15,943&15,945&15,949&15,961&15,993&16,077&16,485&18,291\\
\end{tabular}
}
    \label{tab:memory_requirements}
\end{table*}

\end{document}